\documentclass[conference]{IEEEtran}
\IEEEoverridecommandlockouts
\usepackage{graphicx}
\usepackage{booktabs}
\usepackage{float}
\usepackage{amsmath,amssymb}
\usepackage[hidelinks]{hyperref}
\usepackage{microtype}
\usepackage[htt]{hyphenat}   
\usepackage{titlesec}
\graphicspath{{figures/}}

\titleformat{\section}{\centering\normalfont\normalsize\scshape}{\thesection.}{0.5em}{}
\titlespacing*{\section}{0pt}{0.6ex plus .1ex}{0.4ex plus .1ex}

\title{Beyond Sum-of-Costs: Plan-Space Sampling\\ for Execution-Time Optimization in MAPF}

\author{
\IEEEauthorblockN{Abhishek S\textsuperscript{*}}
\IEEEauthorblockA{\textit{BuildMachineLabs}\\
\href{mailto:abhishekss6363@gmail.com}{mail}}
\and
\IEEEauthorblockN{Badrikanath Praharaj\textsuperscript{*}}
\IEEEauthorblockA{\textit{BuildMachineLabs}\\
\href{mailto:badrikanath.praharaj@gmail.com}{mail}}
\and
\IEEEauthorblockN{Lakshmi Prasad .S \textsuperscript{*}}
\IEEEauthorblockA{\textit{UC Santa Cruz}\\
\href{mailto:lakshmiprasads1818@gmail.com}{mail}}
\and
\IEEEauthorblockN{Sreeram M.V.\textsuperscript{*}}
\IEEEauthorblockA{\textit{BuildMachineLabs}\\
\href{mailto:mvsreeramblr@gmail.com}{mail}}
\thanks{\textsuperscript{*}These authors contributed equally to this work.}
}

\begin{document}
\maketitle

\begin{abstract}
Multi-Agent Path Finding (MAPF) is typically optimized for sum-of-costs (SoC),
yet realized execution time also depends on rotations and ADG-induced blocking.
Optimal rotation-aware planners capture these effects but do not scale; scalable
anytime SoC methods ignore them. We propose Execution-Aware Large Neighborhood
Search (EA-LNS), an anytime repair that re-routes agent neighborhoods against an
analytical ADG completion-time estimate, scoring accepted plans in the SMART
simulator. Across four MovingAI maps, EA-LNS matches an optimal rotation-aware
planner on open maps and significantly surpasses it on congested warehouse
layouts---at agent counts beyond its solvable range---while cutting average
execution time by 19--53\% over an equal-budget MAPF-LNS2 baseline across all
345 paired instances. EA-LNS thus reaches rotation-optimal quality at scale,
exposing analytical-estimate optimism under congestion.
\end{abstract}

\section{Introduction}
Classical MAPF optimizes discrete graph objectives such as sum-of-costs (SoC)
and makespan~\cite{stern2019}, enabling scalable anytime methods including
MAPF-LNS and MAPF-LNS2~\cite{lns2021,mapflns2}. For executed robot teams,
however, SoC is only a partial proxy. SMART is a high-fidelity kinodynamic
simulator for executing MAPF plans~\cite{smart}. SMART-based analyses and the
CADENCE hardware study show that primitive motion burden and ADG/TPG
coordination structure explain execution-time variation beyond discrete path
cost~\cite{yan2025,cadence}.

Execution-aware MAPF spans turn actions, kinodynamic arrival-time objectives,
temporal-plan repair, speed-profile refinement, and learned execution-time
prediction~\cite{zhang2023turn,mass,btpg,winktpg,remap}. Yet a dichotomy
persists: optimal rotation-aware planners (e.g., turn-cost CBS~\cite{zhang2023turn})
minimize execution cost exactly but scale only to small teams, whereas anytime
SoC methods such as MAPF-LNS2 scale yet remain execution-blind. We ask whether
scalable anytime plan-space search can attain---and sustain at scale---the
realized execution quality of an optimal rotation-aware planner from a tractable
analytical model alone.

We introduce Execution-Aware Large Neighborhood Search (EA-LNS), an anytime MAPF
optimizer that targets execution-time quality rather than discrete path cost
alone. EA-LNS is optimized with a tractable ADG-based execution model, while
SMART serves as the common kinodynamic evaluator for the accepted plans.

This paper contributes:
\begin{itemize}
    \item \textbf{C1:} EA-LNS, an execution-aware anytime repair algorithm for
    improving realized MAPF execution time.
    \item \textbf{C2:} across four MovingAI maps, EA-LNS matches an optimal
    rotation-aware planner on open maps, surpasses it significantly on warehouse,
    and scales past where the rotation-aware planner cannot solve.
    \item \textbf{C3:} SMART-measured evidence that execution-aware repair
    improves AET, together with a characterization of analytical-estimate
    optimism under congestion.
\end{itemize}

\section{Method}
\textbf{Execution-time functional.}
Let $\Pi=\{\pi_1,\ldots,\pi_M\}$ be a feasible MAPF plan on a 4-connected grid.
EA-LNS does not treat $\Pi$ merely as a collection of unit-cost paths. Instead,
it maps $\Pi$ to a weighted execution graph whose vertices are primitive robot
actions and whose edges encode causal precedence. Each agent trajectory is
expanded into forward-motion and in-place-rotation primitives. Type-1 edges
preserve the action order of a single agent, while Type-2 edges impose the
planned traversal order at shared workspace locations, yielding an Action
Dependency Graph (ADG)~\cite{honig2016}.

Each primitive action $v$ is assigned a duration $w(v)$ from the matched
kinodynamic model: translational runs follow a constant-acceleration profile,
and heading changes are represented as fixed-rate in-place rotations. For an
agent $i$, its estimated completion time $T_i^{\mathrm{est}}(\Pi)$ is the
longest weighted path in the ADG ending at its terminal action. EA-LNS therefore
scores a plan by its mean estimated completion time,
$J_{\mathrm{EA}}(\Pi)=\tfrac{1}{M}\sum_{i=1}^{M}T_i^{\mathrm{est}}(\Pi)$.
This functional separates the two execution effects suppressed by SoC: primitive
motion burden enters through action weights, and coordination delay enters
through Type-2 dependencies when they lie on critical paths.

\textbf{Execution-aware neighborhood search.}
EA-LNS uses $J_{\mathrm{EA}}$ as the objective inside an anytime
large-neighborhood search. Starting from a feasible incumbent, each iteration
selects a subset of agents, repairs their paths while holding the remaining
paths fixed, and evaluates the resulting candidate by rebuilding its weighted
ADG. A candidate replaces the incumbent only when it improves
$J_{\mathrm{EA}}$. Thus the search remains in discrete MAPF plan space,
but the acceptance criterion is induced by the plan's estimated execution
structure rather than by discrete path length alone.

\textbf{SMART evaluation.}
EA-LNS ranks candidate plans by $J_{\mathrm{EA}}$. After the planning
budget expires, the accepted plan is executed in SMART under a common
kinodynamic configuration, and the resulting simulator AET is the value reported
in the experiments.

\textbf{Baselines.} On identical instances, one SMART configuration, and AET
metric: the \emph{primary} reference is an optimal rotation-aware planner
(turn-cost CBS~\cite{zhang2023turn}); the \emph{secondary} is equal-budget
MAPF-LNS2~\cite{mapflns2}, isolating the gain from $J_{\mathrm{EA}}$ over SoC repair.

\section{Experiments}
\textbf{Setup.} We evaluate on four MovingAI maps spanning open
(\texttt{random-32-32-20}), congested (\texttt{warehouse-10-20-10-2-1}), and
structured (\texttt{room-64-64-8}, \texttt{maze-32-32-4}) layouts, at team sizes
up to $N{=}200$. For each map and agent count both methods run on the same
scenario files (up to $25$ per cell), so each paired comparison shares identical
obstacles, starts, goals, and team size. The metric throughout is SMART-simulated
average execution time (AET). All planners share an equal $90$\,s core-pinned
budget.

\textbf{Pairing and controls.} Every comparison is paired \emph{within} an
instance---a (map, scenario, team-size) triple on which the two planners share
identical obstacles, start/goal assignments, team size, planning budget, SMART
configuration, and AET metric:
\begin{itemize}
  \item \textbf{EA-LNS vs.\ MAPF-LNS2} (the scalable SoC baseline;
  Table~\ref{tab:results}): all four maps at team sizes $N{=}10$ to $200$---%
  $345$ paired instances ($255$ at $N\in\{50,100,200\}$, $90$ at $N\le30$).
  \item \textbf{EA-LNS vs.\ the optimal rotation-aware planner} (turn-cost
  CBS~\cite{zhang2023turn}; Table~\ref{tab:mapft}): $N\le30$ on random, room,
  and maze, and up to $N{=}50$ on warehouse.
\end{itemize}
We report AET reduction vs.\ MAPF-LNS2 and AET gap vs.\ the optimal planner,
both computed within-instance, with Holm-corrected paired Wilcoxon tests.

\begin{table}[t]
\centering
\caption{SMART AET of EA-LNS vs.\ the equal-budget MAPF-LNS2 baseline.
``Reduction'' is the mean within-instance reduction [95\% CI]; every paired
instance improves in every cell ($A_{12}{=}1.00$; exact paired Wilcoxon
$p\le2\times10^{-3}$ per cell, down to $6\times10^{-8}$ for the $n{=}25$ cells).}
\label{tab:results}
\setlength{\tabcolsep}{4pt}
\resizebox{\columnwidth}{!}{%
\begin{tabular}{llrrr}
\toprule
Map & $N$ & MAPF-LNS2 (s) & EA-LNS (s) & Reduction (\%) \\
\midrule
random & 50 & 66.5 & 44.5 & 33.0 [31.6, 34.3] \\
random & 100 & 78.8 & 53.0 & 32.8 [32.0, 33.5] \\
random & 200 & 107.2 & 78.4 & 26.8 [25.7, 27.9] \\
warehouse & 50 & 120.3 & 59.1 & 50.5 [48.8, 52.1] \\
warehouse & 100 & 140.3 & 65.5 & 53.1 [51.8, 54.4] \\
warehouse & 200 & 167.2 & 78.2 & 53.1 [52.1, 54.0] \\
room & 50 & 152.7 & 95.9 & 37.2 [35.6, 38.7] \\
room & 100 & 181.7 & 122.0 & 32.8 [31.2, 34.4] \\
room & 200 & 233.1 & 187.9 & 19.4 [17.9, 20.9] \\
maze & 50 & 132.9 & 77.6 & 41.3 [39.6, 43.1] \\
maze & 100 & 176.4 & 102.1 & 41.7 [39.7, 43.9] \\
maze & 200 & 279.9 & 210.6 & 24.4 [22.1, 27.0] \\
\bottomrule
\end{tabular}}
\end{table}

\begin{table}[t]
\centering
\caption{EA-LNS vs.\ the optimal rotation planner (turn-cost CBS) at the team
sizes where it solves, SMART AET on identical instances. ``$\Delta$'' is the mean
within-instance gap [95\% CI]; positive $=$ EA-LNS reaches \emph{lower} AET than
the optimum. Warehouse is per-cell Holm-significant (pooled $p{=}2.7\times10^{-10}$).}
\label{tab:mapft}
\setlength{\tabcolsep}{4pt}
\resizebox{\columnwidth}{!}{%
\begin{tabular}{llrrr}
\toprule
Map & $N$ & Optimal (s) & EA-LNS (s) & $\Delta$ (\%) \\
\midrule
warehouse & 10 & 47.7 & 46.5 & $+2.5$ [$+1.3$, $+3.9$] \\
warehouse & 20 & 52.9 & 51.4 & $+2.8$ [$+1.8$, $+3.7$] \\
warehouse & 30 & 55.6 & 54.1 & $+2.6$ [$+1.4$, $+3.5$] \\
warehouse & 50 & 61.4 & 59.1 & $+3.6$ [$+3.1$, $+4.3$] \\
random & 20 & 38.4 & 38.6 & $-0.8$ [$-2.6$, $+1.3$] \\
random & 30 & 41.1 & 41.2 & $-0.3$ [$-2.3$, $+1.3$] \\
room & 10 & 84.6 & 83.2 & $+1.8$ [$-0.6$, $+3.7$] \\
room & 20 & 82.9 & 83.0 & $-0.2$ [$-2.9$, $+2.2$] \\
maze & 10 & 55.1 & 56.2 & $-2.1$ [$-3.9$, $-0.4$] \\
maze & 20 & 62.0 & 65.9 & $-6.3$ [$-7.6$, $-5.2$] \\
\bottomrule
\end{tabular}}
\end{table}

\textbf{Main result.} \emph{Vs.\ the optimal rotation planner
(Table~\ref{tab:mapft}).} Where turn-cost CBS solves, EA-LNS reaches its realized
AET on the open and structured maps (random, room: within noise, n.s.\ after Holm) and
significantly \emph{surpasses} it on warehouse---$+3.1\%$ on average ($51/55$
instances, pooled $p{=}2.7\times10^{-10}$); only in the tight maze corridors does
it trail the optimum ($-4.0\%$). EA-LNS thus attains rotation-optimal execution
quality at team sizes the optimal planner cannot solve.
\emph{Vs.\ equal-budget MAPF-LNS2 (Table~\ref{tab:results}).} On the at-scale grid
EA-LNS lowers SMART AET in all $255$ instances, by $19$ to $53\%$ per cell, with
$A_{12}{=}1.00$ and exact paired Wilcoxon $p\le2\times10^{-3}$ per cell ($6\times10^{-8}$
at $n{=}25$); the gain is largest on the congested warehouse ($50$ to $53\%$).

\begin{figure}[H]
\centering
\includegraphics[width=0.96\columnwidth]{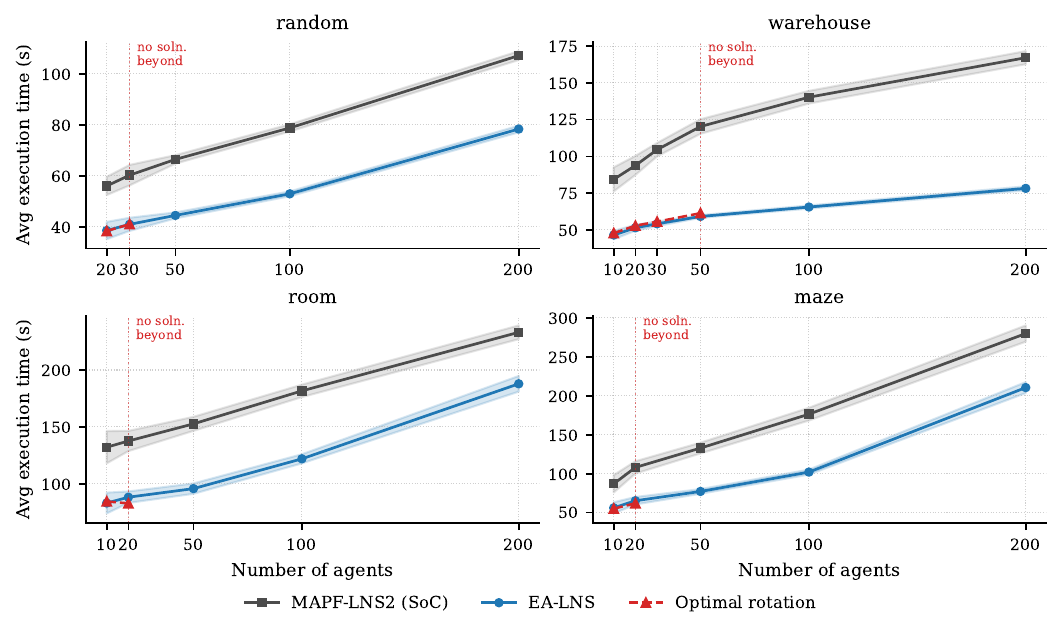}
\caption{SMART AET vs.\ team size (all four maps). EA-LNS (blue) stays below
MAPF-LNS2 (grey); the optimal rotation planner (red) solves only small teams.
Bands: 95\% CIs; lower is better.}
\label{fig:scaling}
\end{figure}

\textbf{Scaling.} Fig.~\ref{fig:scaling} plots absolute AET against team size on
all four maps. EA-LNS stays below the MAPF-LNS2 baseline at every size, while the
optimal rotation planner solves only small teams---producing no solution within
the budget beyond $N\approx30$ ($N\approx50$ on warehouse), where EA-LNS
continues to $N{=}200$. The
EA-LNS--MAPF-LNS2 gap is maintained or grows with scale (warehouse
$\approx$50 to 53\% throughout), softening at $N{=}200$ on the most open and most
constrained maps (random $27\%$, room $19\%$).

\begin{figure}[H]
\centering
\IfFileExists{figures/fig1_anytime.pdf}{%
  \includegraphics[width=0.96\columnwidth]{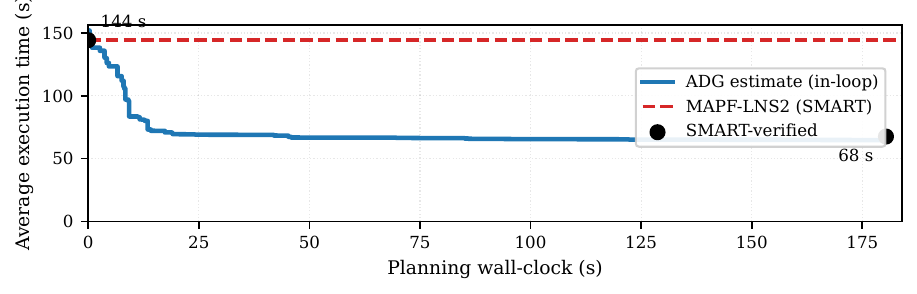}%
}{%
  \fbox{\parbox[c][3.0cm][c]{0.88\columnwidth}{\centering\small
  \emph{[Fig.\ placeholder]} anytime AET vs.\ wall-clock, pending the Titan-B
  trace; render with \texttt{make\_fig1\_anytime.py}.}}%
}
\caption{Anytime convergence (warehouse, $N{=}100$). Estimated AET (blue) falls
over the planning budget, ending well below the equal-budget MAPF-LNS2 baseline
(dashed); both endpoints are SMART-verified.}
\label{fig:anytime}
\end{figure}

\textbf{Anytime convergence.} Fig.~\ref{fig:anytime} shows the repair on a single
representative instance: AET falls monotonically as the sampler accepts improving
neighborhoods, crossing below the equal-budget MAPF-LNS2 baseline and continuing
to improve until the planning budget expires.

\textbf{Mechanism.} EA-LNS targets two execution costs SoC omits: in-place
rotations and inter-agent (Type-2) dependencies in the action graph. Rotation
reduction drives the gains on open maps, while Type-2 relaxation---concentrated
under congestion---drives the warehouse result. A turn-cost-optimal planner
minimizes only its graph objective and cannot reshape this coupling, so EA-LNS
surpasses it on \emph{realized} execution time exactly where coupling, not
rotation, is the binding constraint.

\section{Conclusion and Future Work}
EA-LNS, an anytime plan-space sampler optimizing a tractable analytical
execution-time estimate, matches a turn-cost-optimal planner on open maps and
\emph{surpasses} it on the congested warehouse---by relaxing inter-agent coupling
the optimal planner cannot---while scaling beyond its solvable range and cutting
realized execution time $19$ to $53\%$ over equal-budget MAPF-LNS2. Future work:
a periodic exact-objective check for the estimate's congestion-driven optimism,
and a learned policy targeting high-coupling agents to widen the warehouse gain.

\end{document}